\documentclass[conference]{IEEEtran}
\IEEEoverridecommandlockouts
\usepackage[switch]{lineno}
\usepackage{cite}
\usepackage{amsmath,amssymb,amsfonts}
\usepackage{graphicx}
\usepackage{textcomp}
\usepackage{xcolor}
\usepackage{algorithm}
\usepackage{float} 
\usepackage{soul}
\usepackage{caption}
\usepackage{subcaption}
\usepackage{algpseudocode}
\usepackage{cite}
\usepackage[hidelinks]{hyperref}
\def\BibTeX{{\rm B\kern-.05em{\sc i\kern-.025em b}\kern-.08em
    T\kern-.1667em\lower.7ex\hbox{E}\kern-.125emX}}
\begin{document}

\title{Explainable AI for Smart Greenhouse Control: Interpretability of Temporal Fusion Transformer in the Internet of Robotic Things\\

}

\author{\IEEEauthorblockN{1\textsuperscript{st} Muhammad Jawad Bashir}
\IEEEauthorblockA{\textit{Department of Computing} \\
\textit{Atlantic Technological University}\\
Donegal, Ireland \\
jawad.bashir@research.atu.ie}
\and
\IEEEauthorblockN{2\textsuperscript{nd} Shagufta Henna}
\IEEEauthorblockA{\textit{Department of Computing} \\
\textit{Atlantic Technological University}\\
Donegal, Ireland \\
0000-0002-8753-5467}
\and
\IEEEauthorblockN{3\textsuperscript{rd} Eoghan Furey}
\IEEEauthorblockA{\textit{Department of Computing} \\
\textit{Atlantic Technological University}\\
Donegal, Ireland \\
eoghan.furey@atu.ie}
}

\maketitle

\begin{abstract}
The integration of the Internet of Robotic Things (IoRT) in smart greenhouses has revolutionised precision agriculture by enabling efficient and autonomous environmental control. However, existing time series forecasting models in such setups often operate as black boxes, lacking mechanisms for explainable decision-making, which is a critical limitation when trust, transparency, and regulatory compliance are paramount in smart farming practices. This study leverages the Temporal Fusion Transformer (TFT) model to automate actuator settings for optimal greenhouse management. To enhance interpretability and trust in the model decision-making process, both local and global explanation techniques were employed using model-inherent interpretation, local interpretable model-agnostic explanations (LIME), and SHapley additive explanations (SHAP). These explainability methods provide information on how different sensor readings, such as temperature, humidity, $CO_2$ levels, light, and outer climate, contribute to actuator control decisions in an automated greenhouse. The trained TFT model achieved a test accuracy of 95\% on a class-imbalanced dataset for actuator control settings in an automated greenhouse environment. The results demonstrate the varying influence of each sensor on real-time greenhouse adjustments, ensuring transparency and enabling adaptive fine-tuning for improved crop yield and resource efficiency.
\end{abstract}


\section{Introduction}

The world population is estimated to increase to 9.8 billion by 2050 and to reach 10.4 billion by 2084 \cite{owid-un-population-2024-revision}. According to a UN report in 2024, hunger is still one of the key challenges in the world; 281.6 million people suffered acute food insecurity in 2023, projected to more than double (582 million) in the next 5 years \cite{UN2024}. The dynamics of climate change, combined with limited resources, threaten food systems, leading to unstable food prices and availability. Based on the OECD-FAO Agricultural Outlook report for 2023 to 2032, crop production growth requires investment in improving yields and advancing farm management practices \cite{oecdOECDFAOAgriculturalOutlook2023}. Given this situation, global agricultural practices must be revised to promote resilience and sustainability while maintaining efficient resource usage and maximum yield. Integrating the Internet of Robotic Things (IoRT) in precision agriculture has significant potential to address these challenges by enabling autonomous and efficient environmental control.

IoRT-based precision agriculture applications increasingly use actuators, autonomous monitoring vehicles, and intelligent greenhouse control systems to optimise various agricultural practices, such as irrigation scheduling, harvesting, and climate regulation \cite{wakchaureApplicationAITechniques2023}. Several deep learning approaches have recently been proposed to automate greenhouse control and operation with increased productivity while reducing operational costs. For example, \cite{zhaoParallelControlGreenhouse2022} introduced a transfer learning-based LSTM model for greenhouse climate forecasting, achieving improved temperature, humidity, and $CO_2$ levels predictions using simulated and real-world data. \cite{ullahOptimizationSchemeIoT2022} developed an IoT-driven climate control system that integrated Kalman filtering with actuator control to optimise environmental conditions, resulting in a 26.56\% reduction in energy consumption. \cite{liForecastingGreenhouseAir2024a} further enhanced forecasting accuracy using an attention-based LSTM with wavelet-based denoising for climate and soil temperature predictions, achieving $R^2$ scores above 0.93. In addition to climate regulation, \cite{guoMultimodelFusionMethod2024} proposed an ensemble LSTM-attention model for $CO_2$ monitoring, while \cite{daiVTNetMultidomainInformation2024} employed multi-modal learning with frequency-domain analysis and genetic algorithms for long-term water usage prediction, achieving up to 46.93\% improvement in irrigation accuracy. Yield forecasting applications have also seen promising advancements. \cite{demirhanDeepLearningFramework2025} utilized deep neural networks with long-term climatic data for multi-crop yield prediction, and \cite{liuAIdrivenTimeSeries2025} proposed a multi-modal LSTM-TCN framework for the prediction of weekly strawberry yield. Despite their promising outcomes, many of these approaches lack transparency and generalizability, highlighting the need for interpretable models that can support trustworthy and sustainable greenhouse automation.

Explainable Artificial Intelligence (XAI) is increasingly being incorporated into smart agriculture to enhance transparency, mitigate decision biases, and foster trust in autonomous systems. Within this domain, greenhouse automation represents a prominent use case of the Internet of Robotic Things (IoRT), where interconnected sensing, actuation, and intelligence converge to enable precise, data-driven environmental control. In such high-stakes applications, ranging from irrigation and ventilation management to crop growth optimisation, explainability allows stakeholders to interpret model behaviour, understand the role of key variables, and ensure fairness and accountability in automated decisions \cite{ryoExplainableArtificialIntelligence2022}.

Current XAI methods for time-series data fall broadly into three categories: attention-based, gradient-based, and model-agnostic approaches. Attention-based architectures such as the Temporal Fusion Transformer (TFT) \cite{TFT}, mixture attention LSTMs \cite{xuInterpretableLSTMBased2022}, and perturbation-driven self-attention models embed interpretability within their design, but remain tied to specific model structures. Gradient-based methods, including Grad-CAM \cite{selvarajuGradCAMVisualExplanations2020} and integrated gradients, are useful for visualising feature salience but require access to internal gradients. Model-agnostic techniques such as SHAP (SHapley Additive exPlanations) \cite{SHAP}, TimeSHAP \cite{bentoTimeSHAPExplainingRecurrent2021}, and LIME (Local Interpretable Model-Agnostic Explanations) \cite{LIME} offer flexibility across architectures but can be computationally intensive and sometimes unstable under high-dimensional inputs.

These techniques have been applied in several agricultural domains, including XAI-driven crop recommendation systems \cite{nagasrinivasuXAIdrivenModelCrop2024, akkemStreamlitbasedEnhancingCrop2024}, grass growth modelling \cite{kennyPredictingGrassGrowth2019}, plant disease detection \cite{plantdiseasexai} and digital twin integration \cite{xaidigitaltwin}. However, in greenhouse automation, where multiple actuators interact in response to complex environmental conditions, existing XAI applications often lack comprehensive global and local interpretability. Moreover, there remains limited work comparing model-inherent and model-agnostic explanations within unified, high-resolution time-series prediction frameworks.

Although transformer-based models have shown strong performance in sequential forecasting tasks within agriculture, they often function as black boxes, offering limited transparency into how sensor readings influence control actions. In autonomous greenhouse environments, where actuators govern critical variables such as temperature, humidity, and radiation, this lack of interpretability impedes stakeholder trust and undermines the collaborative tuning of control strategies. Ensuring transparency in such systems is essential for diagnosing unexpected behaviour, detecting bias, and enabling domain experts to make informed adjustments.

To address these limitations, this study presents an interpretable deep learning framework for IoRT-enabled greenhouse automation. The approach utilizes the Temporal Fusion Transformer (TFT) to forecast actuator states from multivariate sensor inputs. TFT’s built-in Variable Selection Network (VSN) and multi-head attention mechanism provide global interpretability by identifying the most relevant features and time steps. To extend this interpretability beyond model-intrinsic mechanisms, the framework incorporates post-hoc, model-agnostic methods, SHAP and LIME, to derive both global and class-specific explanations of how sensor dynamics drive actuator decisions.

The model achieves a classification accuracy of 95\% in predicting actuator settings, indicating strong predictive reliability. XAI outputs reveal how environmental parameters, such as air temperature, humidity, CO\textsubscript{2} concentration, light intensity, and soil moisture, contribute to real-time control actions. These insights enable transparent decision-making, facilitate adaptive fine-tuning, and support the development of robust, interpretable greenhouse control systems within broader IoRT-based precision agriculture frameworks.

\section{Network Model}

\begin{figure}[h]
    \centering
    \includegraphics[width=1\linewidth]{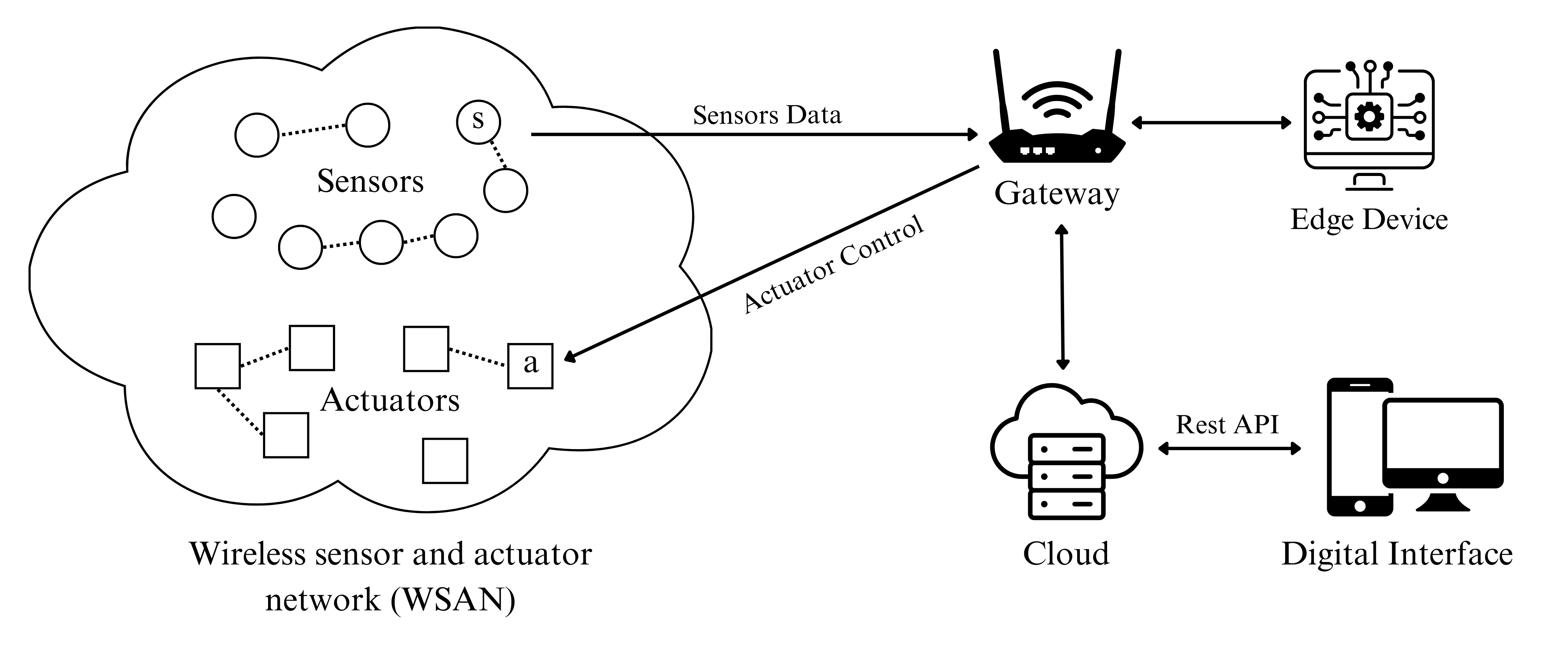}
    \caption{Network model: representing sensors, actuators, gateways and edge devices for greenhouse control.}
    \label{fig:network-model}
\end{figure}

In an IoRT system, the network model consists of three main components: sensors, a gateway, and actuators, all communicating to enable real-time control, as shown in Figure \ref{fig:network-model}. Sensors \( S = \{s_1, s_2, \dots, s_m\} \) collect environmental or system state data, transmitting their measurements \(Y_t = \{ y_t^1, y_t^2, \dots, y_t^m \}\) to the gateway through a wired or wireless network. The gateway \( G \) serves as the intermediary, aggregating sensor data and applying a control function using a deep learning approach on an edge device to predict the optimal interpretable actuator commands. The predicted control signal is then transmitted to the actuators \( A = \{a_1, a_2, \dots, a_n\} \), which execute the required actions. Communication between sensors and the gateway introduces a transmission delay \( d_s \), while communication between the gateway and actuators introduces a delay in actuation \( d_a \).

The network is constrained by factors such as bandwidth limitations, energy consumption, latency, and packet loss, all of which affect data transmission and control accuracy. The objective of the IoRT network is to efficiently map sensor measurements to actuator commands while minimising actuator energy consumption, subject to the latency constraint.

\section{Problem Formulation}
In an IoRT system for greenhouse control, the objective is to control a set of actuators \( A = \{a_1, a_2, \dots, a_n\} \) using sensor measurements from the set of sensors $S$. The state of the system at time \( t \) is represented by \( x_t \in \mathbb{R}^d \), while the sensors provide measurements. 

\begin{equation} \label{eq2}
y_t = h(x_t)
\end{equation}

\noindent
where \( y_t \) represents the sensor outputs at time \( t \). The actuators are controlled by an action vector \( u_t \in \mathbb{R}^n \), which is determined by a control policy that maps sensor measurements to actuator commands:  

\begin{equation} \label{eq3}
u_t = g(y_t).
\end{equation}

The objective is to find the optimal control policy that minimizes the actuator energy consumption over a finite time horizon \( T \), subject to sensor measurements and control constraints:

\begin{equation} \label{eq4}
\min_{u_t} \sum_{t=0}^{T} \sum_{i=1}^{n} c_i u_{i,t}^2
\end{equation}

Here, \( c_i \) are the cost coefficients that represent the energy consumption of each actuator. The goal is to determine the optimal control actions \( u_t \) that minimise the total energy cost while ensuring appropriate actuator control based on sensor measurements.

\section{Proposed Explainable AI-enabled TFT Framework}
This section presents the XAI-driven deep learning framework to predict greenhouse control settings using sensor readings in an IoRT-enabled greenhouse deployment. The proposed framework incorporates TFT \cite{TFT} coupled with local and global explanations to improve transparency and interpretability. The XAI module consists of a Variable Selection Network to assess the importance of features, combined with SHAP \cite{SHAP} for global explanation and LIME \cite{LIME} for local explanation. The main goal is to improve the transparency and accuracy of decisions made using the IoRT-enabled AI framework in precision agriculture. 

\begin{figure*}[h]
    \centering
    \includegraphics[width=1\linewidth]{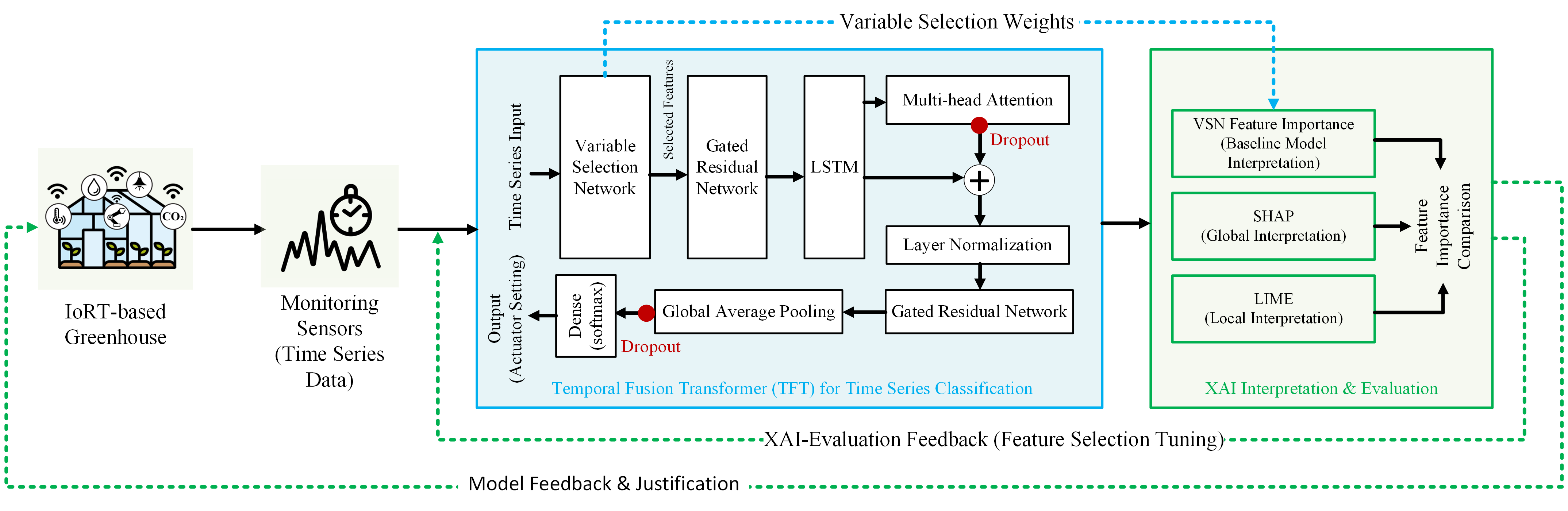}
    \caption{XAI-based greenhouse automation framework: IoRT-enabled monitoring, TFT for classification, and XAI(VSN, SHAP, LIME) for explanation of predictions.}
    \label{fig:approach}
\end{figure*}

The proposed framework, as shown in Figure \ref{fig:approach} consists of four key components, including data input, learning of temporal dependencies, and sequential patterns via TFT and XAI-based feature selection using VSN-based importance score, SHAP, and LIME. These components are briefly described below.

\begin{algorithm}[h]
\caption{XAI-based TFT greenhouse control.}
\label{alg:tft_xai}
\begin{algorithmic}[1]


\State \textbf{Input:} Sensor data $\mathbf{S} \in \mathbb{R}^{N \times F}$, dropout rate $p$,  batch size $B$, number of classes $K$
                       Actuator labels $\mathbf{y} \in \{0,\ldots,K{-}1\}^{N}$

\State \textbf{Output:} Trained TFT model $\mathcal{M}$, 
        updated feature importance $\mathbf{I}$ (via LIME/SHAP/VSN)

\Statex \textbf{Part -1: Classification Model}
\State \textbf{(i)Data Preparation}
\State \quad $\mathbf{X} \gets \Call{Preprocess}{\mathbf{S}}$ 
            \Comment{e.g., fill NAs, scale features}
\State \quad $(\mathbf{X}, \mathbf{y}) \gets \Call{GenerateRollingWindow}{\mathbf{X}, \mathbf{y}, w}$
            \Comment{shape: $(N-w+1, w, F)$}
\State \quad $(\mathbf{X}, \mathbf{y}) \gets \Call{ScaleData}{\mathbf{X}, \mathbf{y}}$

\Statex
\State \textbf{(ii) TFT}
\State \quad $\mathcal{V} \gets \Call{VariableSelectionNetwork}{d_{\mathrm{model}}, p}$
\State \quad $\mathcal{L} \gets \Call{LSTMEncoder}{d_{\mathrm{model}}}$
\State \quad $\mathcal{A} \gets \Call{MultiHeadAttention}{H, d_{\mathrm{model}}}$
\State \quad $\mathcal{O} \gets \Call{GlobalPoolAndDense}{K}$
\State \quad $\mathcal{M} \gets \Call{BuildTFTModel}{\mathcal{V}, \mathcal{L}, \mathcal{A}, \mathcal{O}}$

\Statex
\State \textbf{(iii) Training and Evaluation}
\State \quad $(\mathbf{X}_{\mathrm{train}}, \mathbf{y}_{\mathrm{train}}), 
                  (\mathbf{X}_{\mathrm{test}}, \mathbf{y}_{\mathrm{test}}) 
                  \gets \Call{SplitData}{\mathbf{X}, \mathbf{y}}$
\State \quad \textbf{for} $e = 1 \ldots E$ \textbf{do}
\State \quad \quad $\Call{TrainBatch}{\mathcal{M}, \mathbf{X}_{\mathrm{train}}, \mathbf{y}_{\mathrm{train}}, B}$
\State \quad \textbf{end for}
\State \quad $(\hat{\mathbf{y}}, \mathbf{I}_{\mathrm{VSN}}) \gets \Call{Predict}{\mathcal{M}, \mathbf{X}_{\mathrm{test}}}$
\State \quad $\Call{Evaluate}{\hat{\mathbf{y}}, \mathbf{y}_{\mathrm{test}}}$
            \Comment{e.g., accuracy, confusion matrix}

\Statex \textbf{Part -2: XAI Evaluation \& Feature Importance Retuning}
\State \quad $E_{\mathrm{LIME}} \gets \Call{LIMEExplain}{\mathcal{M}, \mathbf{X}_{\mathrm{train}}, \mathbf{X}_{\mathrm{test}}}$
\State \quad $E_{\mathrm{SHAP}} \gets \Call{SHAPExplain}{\mathcal{M}, \mathbf{X}_{\mathrm{train}}, \mathbf{X}_{\mathrm{test}}}$
\State \quad $\mathbf{I} \gets \Call{RetuneFeatureImportance}{\mathbf{I}_{\mathrm{VSN}}, E_{\mathrm{LIME}}, E_{\mathrm{SHAP}}}$
\State \quad $\Call{FineTuneFeatureSelection}{\mathcal{M}, \mathbf{I}}$ 
\State \quad \textbf{return} $(\mathcal{M}, \mathbf{I})$

\end{algorithmic}
\end{algorithm}

\subsection{Temporal Fusion Transformer}
TFT was initially proposed for multi-horizon time series forecasting in an interpretable fashion. It integrates a recurrent neural network, the gating mechanism, and multi-head attention to capture temporal features from both static and non-static variables. The model-based interpretability \cite{aliExplainableArtificialIntelligence2023} is incorporated through the variable selection network enhanced with multi-head interpretable attention. These components collectively enable TFT to extract and quantify the influential features, facilitating deeper insight into the forecasting process.

The VSN adaptively selects input features per time step using a gated residual network (GRN), weighting variables based on contextual relevance. It effectively reduces dimensionality by retaining critical features and discarding irrelevant ones, enhancing the transformer's robustness for large, complex, real-time datasets. GRN leverages residual connections to efficiently model non-linear relationships with fewer parameters, improving interpretability and generalization. The gating mechanism manages information flow and retains essential long-term and short-term dependencies, accelerating learning by mitigating vanishing gradients and enhancing model performance. Long Short-Term Memory (LSTM) captures temporal dependencies, long-term trends, and seasonal patterns. However, inherently less interpretable, integration within TFT enhances transparency. LSTM-generated temporal embeddings feed into multi-head attention, addressing local temporal dependencies and strengthening sequential bias handling, improving model robustness across diverse datasets. TFT also employs an interpretable multi-head attention mechanism with shared value representations across heads. This design improves the aggregation of attention weights, enhancing representational power and interpretability, allowing intuitive visualization of temporal feature importance.

\subsection{Explainability of Temporal Fusion Transformer}

LIME, introduced by \cite{LIME}, explains the predictions of a machine learning approach, with a specific focus on black-box models, such as neural networks. LIME advocates for the local explanation of the model by approximating it as a decision tree to capture the local relationships among the data points. To capture a local explanation of the model, LIME generates perturbations based on a sample to observe its impact on the predictions. Specifically, it compares the behavior of the model under both the perturbed and original samples. The resultant coefficients of the explanation of the original and perturbed sample are then interpreted as the explanation of the model prediction. LIME offers insights into the relevance of the features to a particular greenhouse control prediction, making it suitable for greenhouse operation where interpretability is crucial for planning and deployment of actuators and sensors, optimizing productivity and operational cost, while reducing the environmental toll of large-scale greenhouses.

SHAP \cite{SHAP} employs a game-theoretical approach to explanation, where Shapley values are calculated based on the fair payout among the features for their contribution to predictions. The contribution towards the prediction is computed based on the different feature permutations, making SHAP a robust approach towards explanation. In contrast to LIME, SHAP provides both global and local explanations, offering a locally accurate explanation of the predictions. In the context of the greenhouse, this is particularly important to understand, globally, a summary of all crucial features that represent varying sensor readings over time, and locally, a breakdown of individual predictions, i.e., greenhouse controls, into a sum of all features, i.e., sensor readings.

The algorithm~\ref{alg:tft_xai} outlines the pipeline for actuator prediction with interpretability. Sensor readings are first pre-processed and labelled (lines 3–6), then passed to an TFT model built using a variable selection network, LSTM encoder and multi-head attention (lines 7–11). The model captures relevant features, temporal dependencies, and global context to form a robust predictive architecture. Training and evaluation follow in lines 13–17, where the dataset is split, the model is trained, and predictions are evaluated using accuracy and a confusion matrix. Lines 18–22 introduce the explainability module: LIME and SHAP are applied to compute local and global feature importance, which are then aggregated to refine the relevance of features. The model may optionally be fine-tuned based on these insights before returning final predictions and interpretability outputs.

\section{Results \& evaluation}
\subsection{Dataset \& preprocessing}

The dataset used in this investigation comes from the Autonomous Greenhouse Challenge \cite{dataset_2}, aimed at automating cherry tomato cultivation in greenhouse environments. It consists of continuous monitoring and optimal control data over six months, recorded at uniform intervals of five minutes, involving measurements from 28 sensors and control actions from 16 actuators. Sensors used for environmental monitoring include devices that capture external parameters such as air temperature, relative humidity, solar radiation, wind speed, rainfall, photosynthetically active radiation (PAR) and absolute humidity. In the greenhouse, sensor types include aspirated temperature sensors, relative humidity sensors, gas sensors (for $CO_2$ measurement), pipe temperature sensors, and irrigation monitoring sensors. The actuators manage critical environmental control points including heating, ventilation, $CO_2$ dosing, irrigation schedules, assimilation lighting (HPS and LED lamps), and curtain positioning systems, ensuring optimal greenhouse climate conditions for plant growth.

\begin{table}[htbp]
    \centering
    \renewcommand{\arraystretch}{1.2}
    \begin{tabular}{|c|p{6cm}|}
        \hline
        \textbf{Setting Class} & \textbf{Actuators} \\
        \hline
        Class 1 & \texttt{t\_rail\_min\_vip}, \texttt{t\_ventwind\_vip} \\
        \hline
        Class 2 & \texttt{assim\_vip}, \texttt{int\_farred\_vip}, \texttt{int\_white\_vip} \\
        \hline
        Class 3 & \texttt{co2\_vip}, \texttt{int\_red\_vip}, \texttt{scr\_enrg\_vip}, \texttt{window\_pos\_lee\_vip} \\
        \hline
        Class 4 & \texttt{dx\_vip}, \texttt{t\_heat\_vip}, \texttt{water\_sup\_intervals\_vip\_min}, \texttt{t\_ventlee\_vip} \\
        \hline
        Class 5 & \texttt{scr\_blck\_vip} \\
        \hline
        Class 6 & \texttt{int\_blue\_vip}, \texttt{t\_grow\_min\_vip} \\
        \hline
    \end{tabular}
    \caption{Actuators grouped by setting class}
    \label{tab:setting_classes}
\end{table}

\begin{figure*}[t]
    \centering
    \begin{subfigure}{0.32\textwidth}
        \centering
        \includegraphics[width=\linewidth]{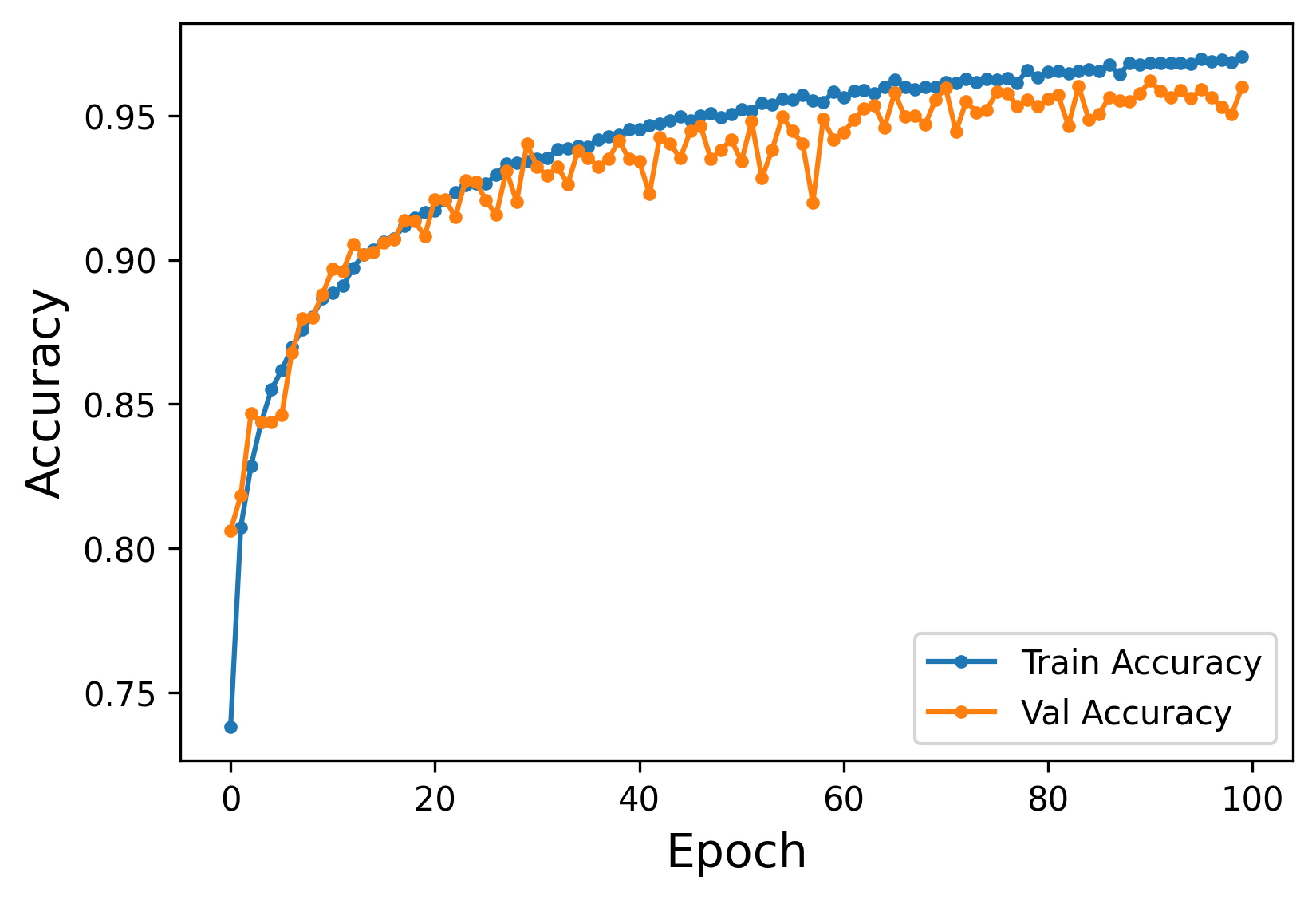}
        \subcaption{}
        \label{fig:training_accuracy}
    \end{subfigure}
    \hfill
    \begin{subfigure}{0.32\textwidth}
        \centering
        \includegraphics[width=\linewidth]{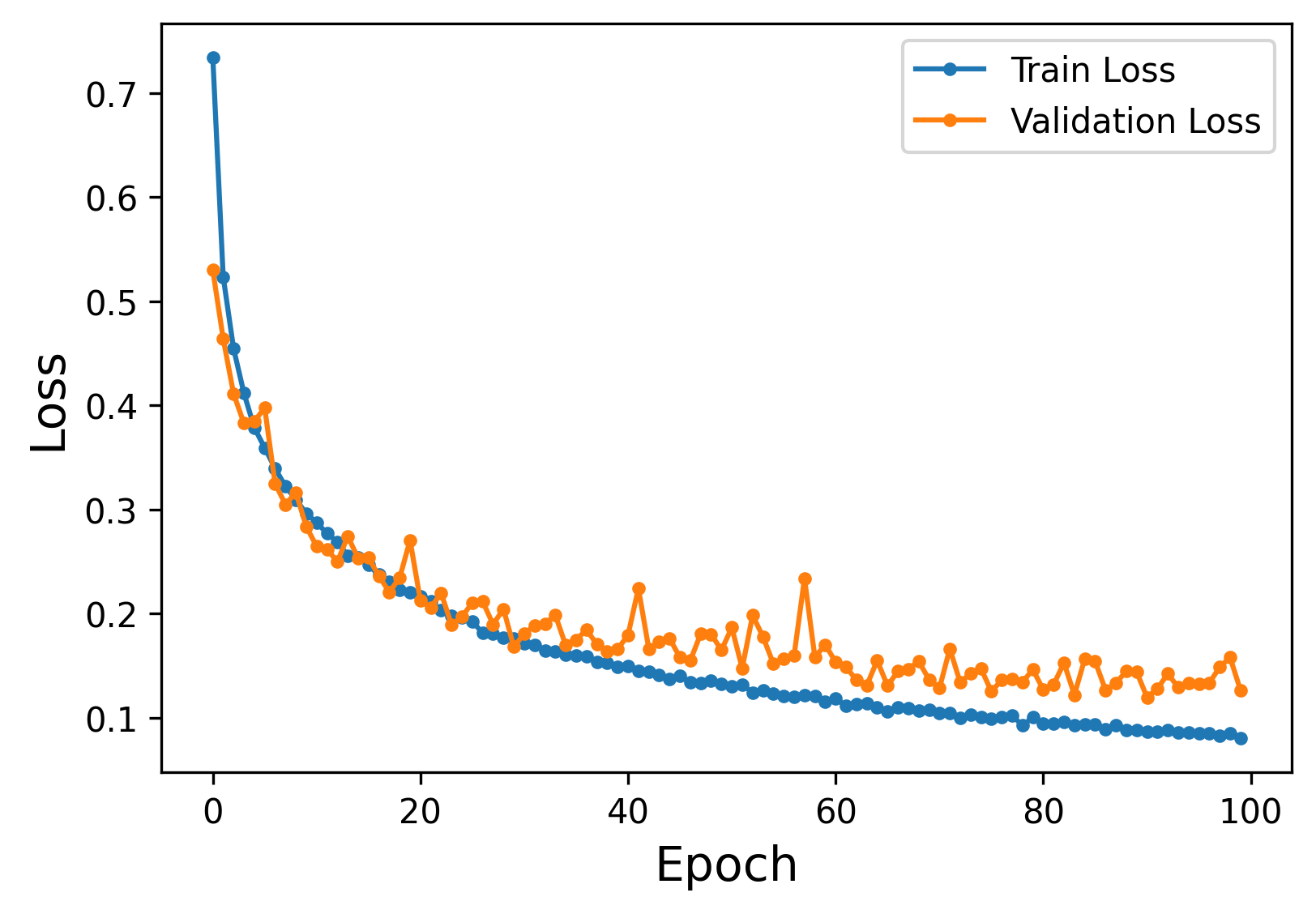}
        \subcaption{}
        \label{fig:training_loss}
    \end{subfigure}
    \hfill
    \begin{subfigure}{0.32\textwidth}
        \centering
        \includegraphics[width=\linewidth]{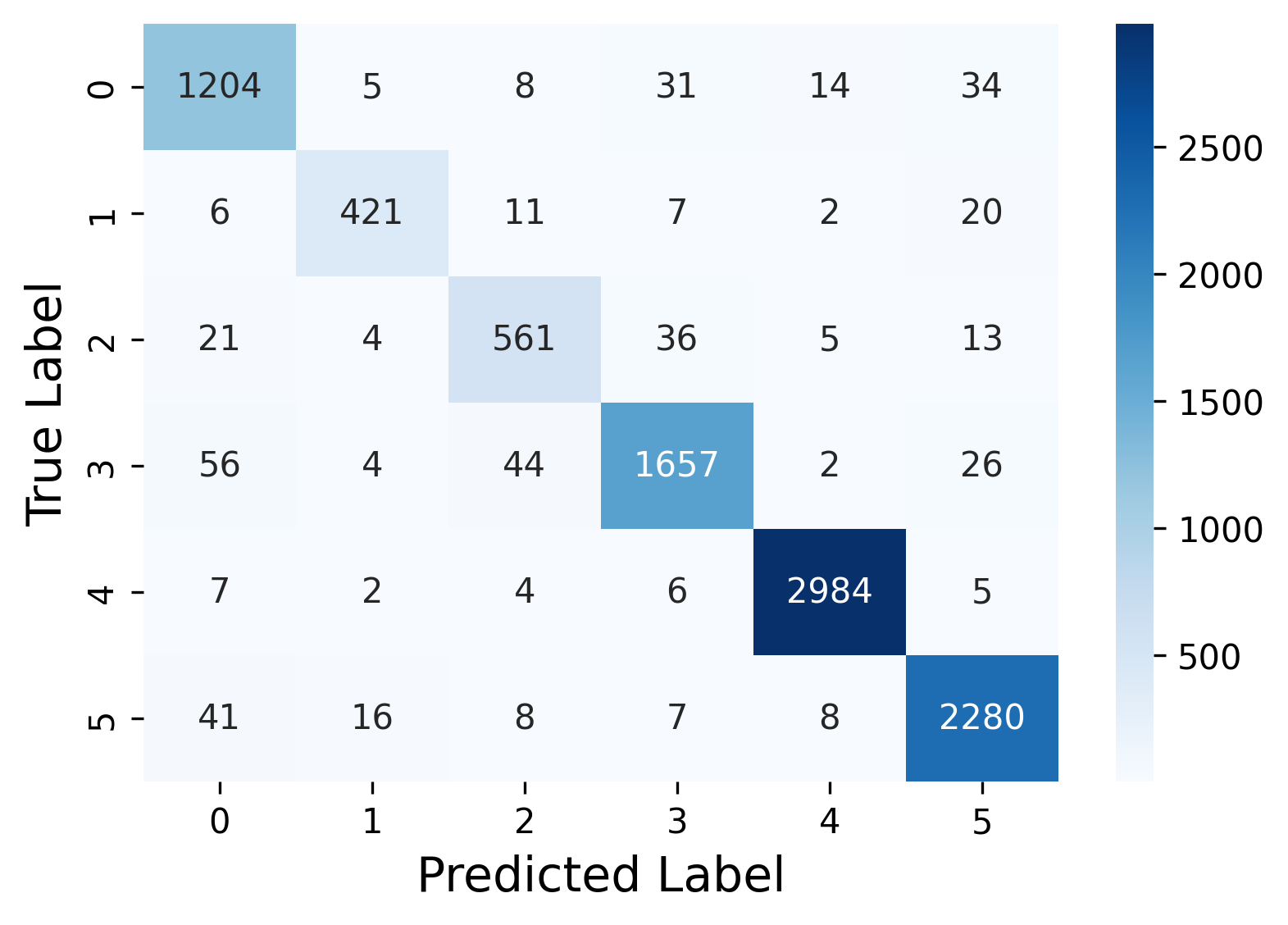}
        \subcaption{}
        \label{fig:confusion_matrix}
    \end{subfigure}
    \caption{Model training performance and evaluation results: (a) accuracy, (b) loss, (c) confusion matrix.}
    \label{fig:model_performance}
\end{figure*}

\begin{figure*}[htbp]
    \centering
    \begin{minipage}{0.45\textwidth}
        \centering
        \includegraphics[width=\linewidth]{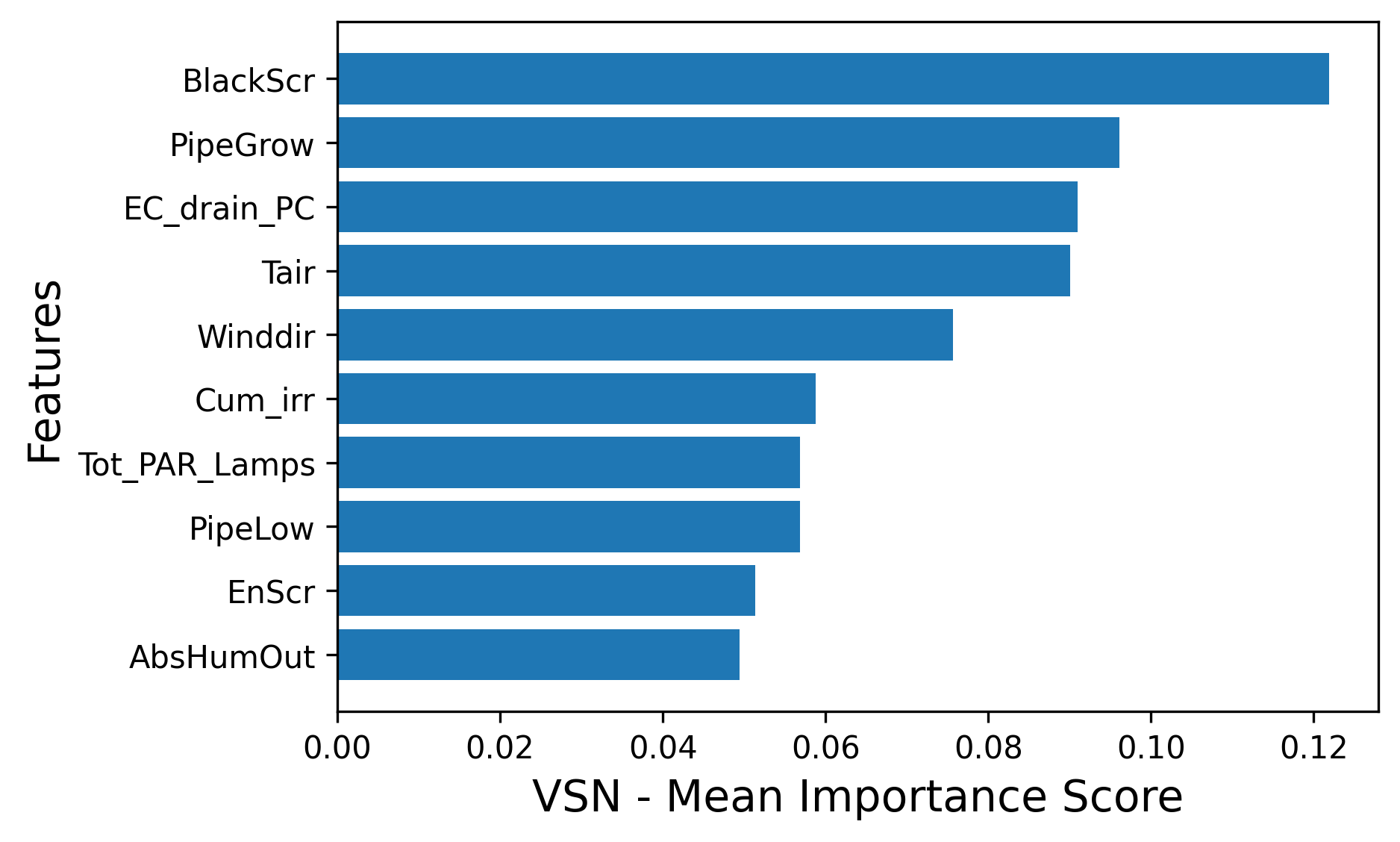}
        \subcaption{}
        \label{fig:vsn}
    \end{minipage}
    \hfill
    \begin{minipage}{0.45\textwidth}
        \centering
        \includegraphics[width=\linewidth]{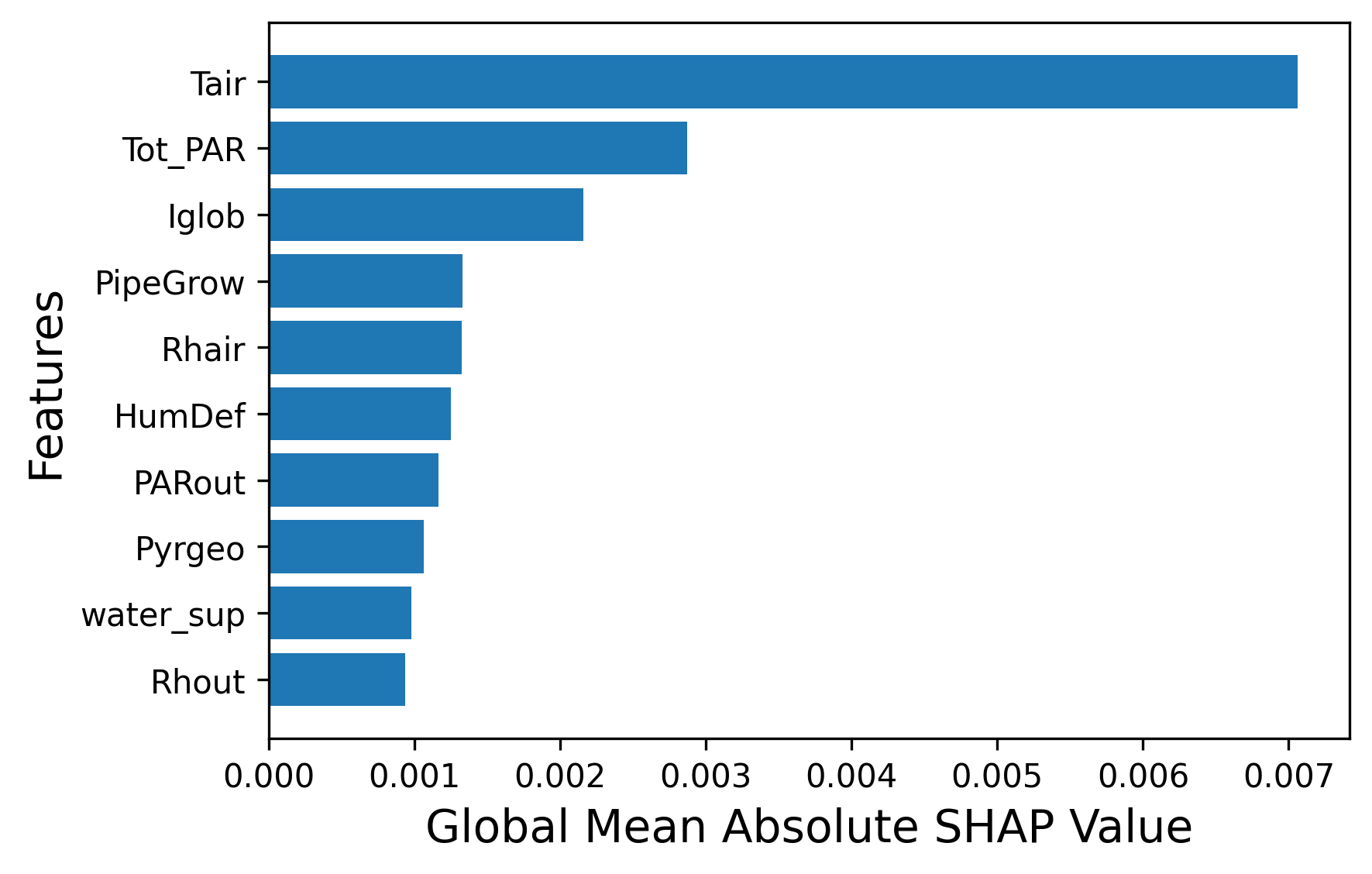}
        \subcaption{}
        \label{fig:shap}
    \end{minipage}
    \caption{Global feature importance based on VSN (left) and SHAP (right): top-10 contributing features.}
    \label{fig:shap_vsn_explanation}
\end{figure*}

\begin{figure*}[htbp]
    \centering
    \begin{minipage}{0.32\textwidth}
        \centering
        \includegraphics[width=\linewidth, trim={0 0 0 0.8cm}, clip]{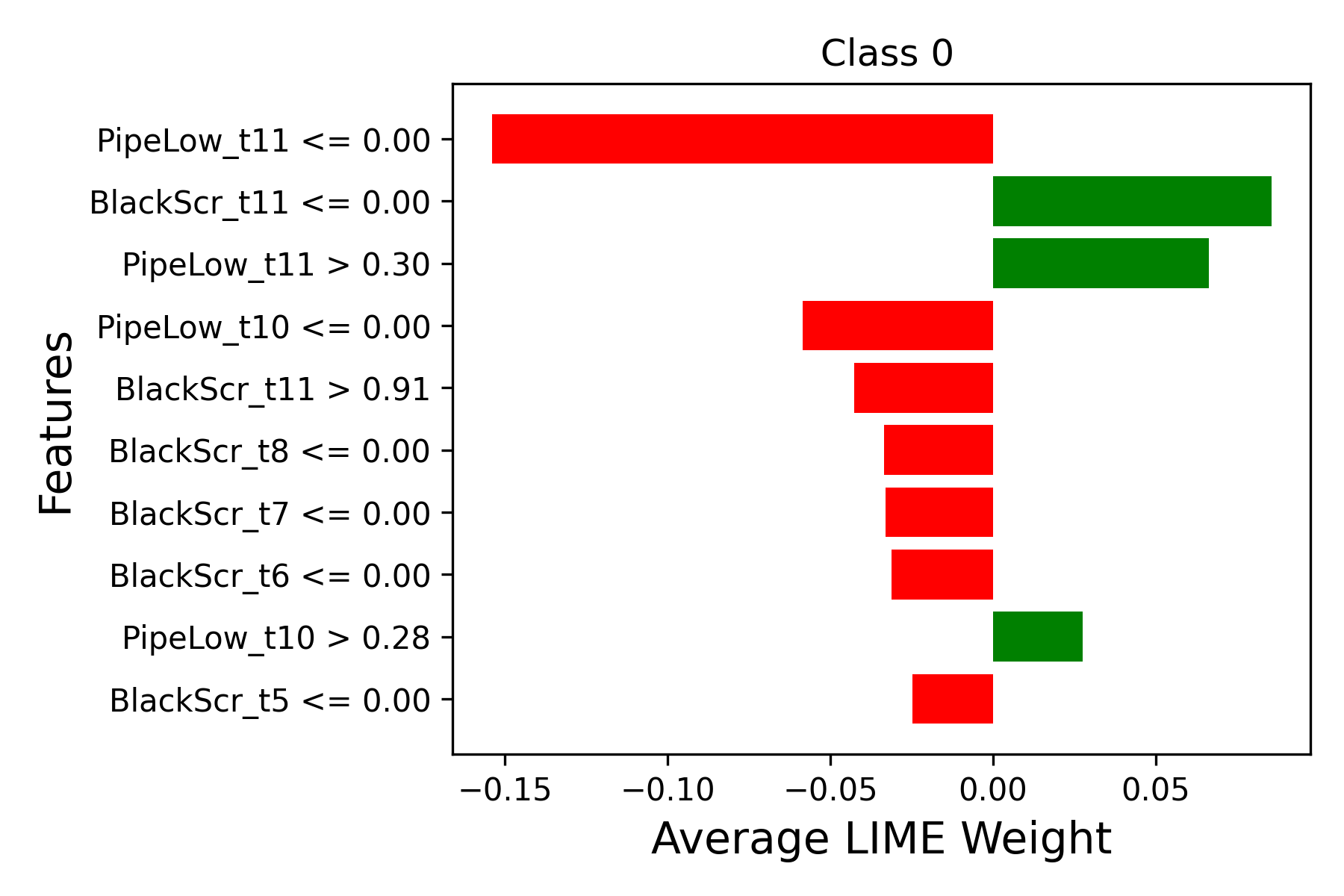}
        \subcaption{}
    \end{minipage}
    \hfill
    \begin{minipage}{0.32\textwidth}
        \centering
        \includegraphics[width=\linewidth, trim={0 0 0 0.8cm}, clip]{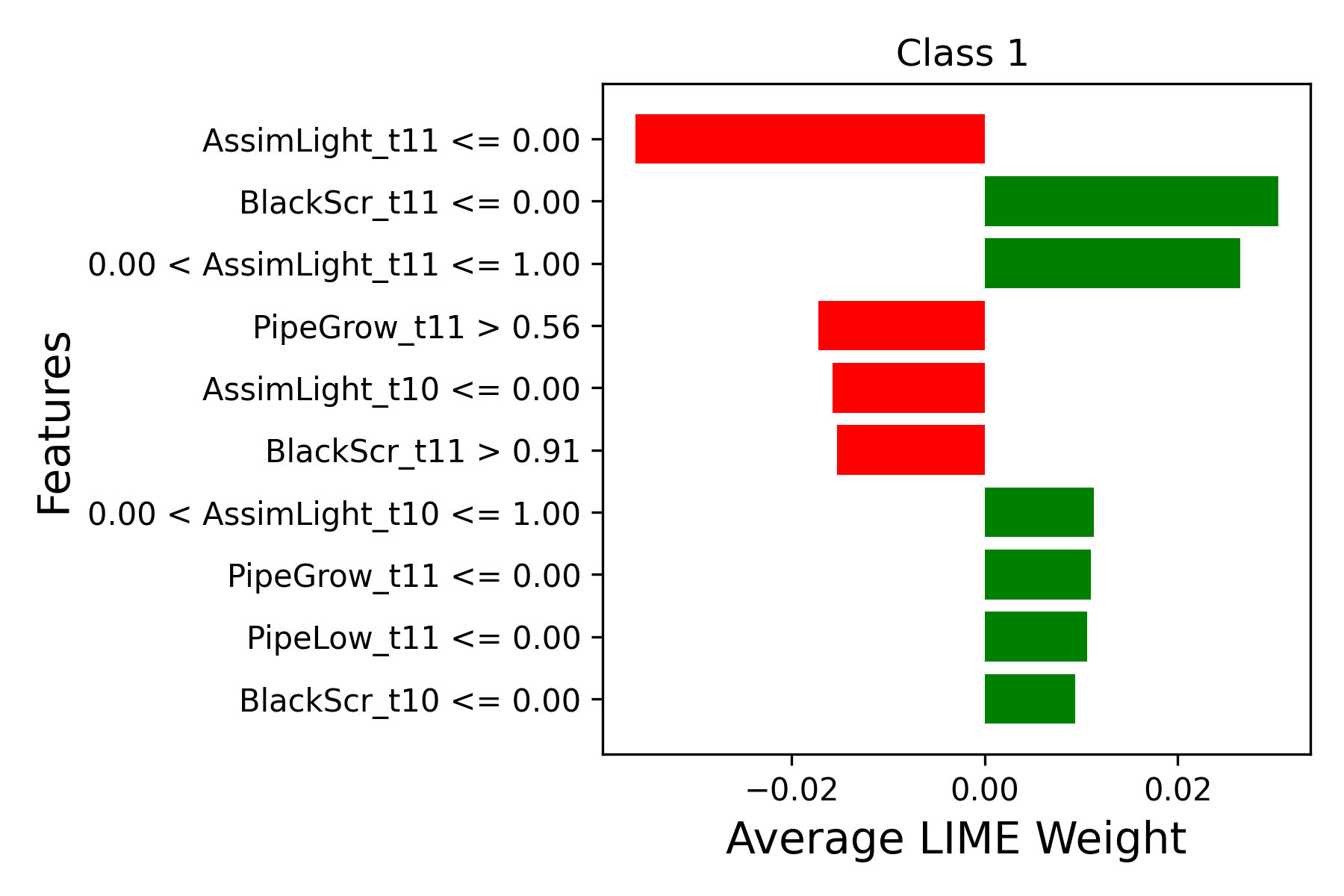}
        \subcaption{}
    \end{minipage}
    \hfill
    \begin{minipage}{0.32\textwidth}
        \centering
        \includegraphics[width=\linewidth, trim={0 0 0 0.8cm}, clip]{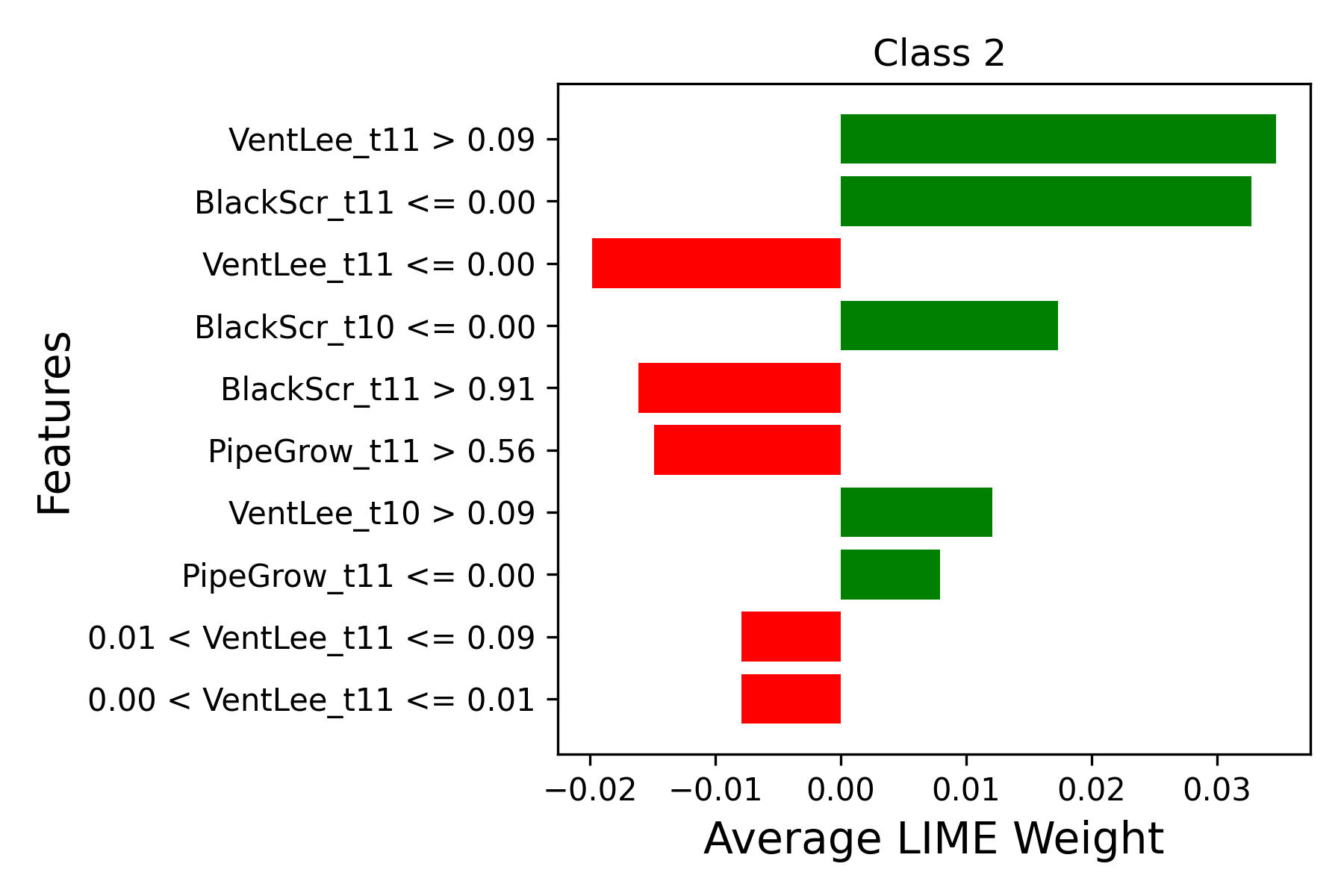}
        \subcaption{}
    \end{minipage}
    
    \vspace{1mm}
    
    \begin{minipage}{0.32\textwidth}
        \centering
        \includegraphics[width=\linewidth, trim={0 0 0 0.8cm}, clip]{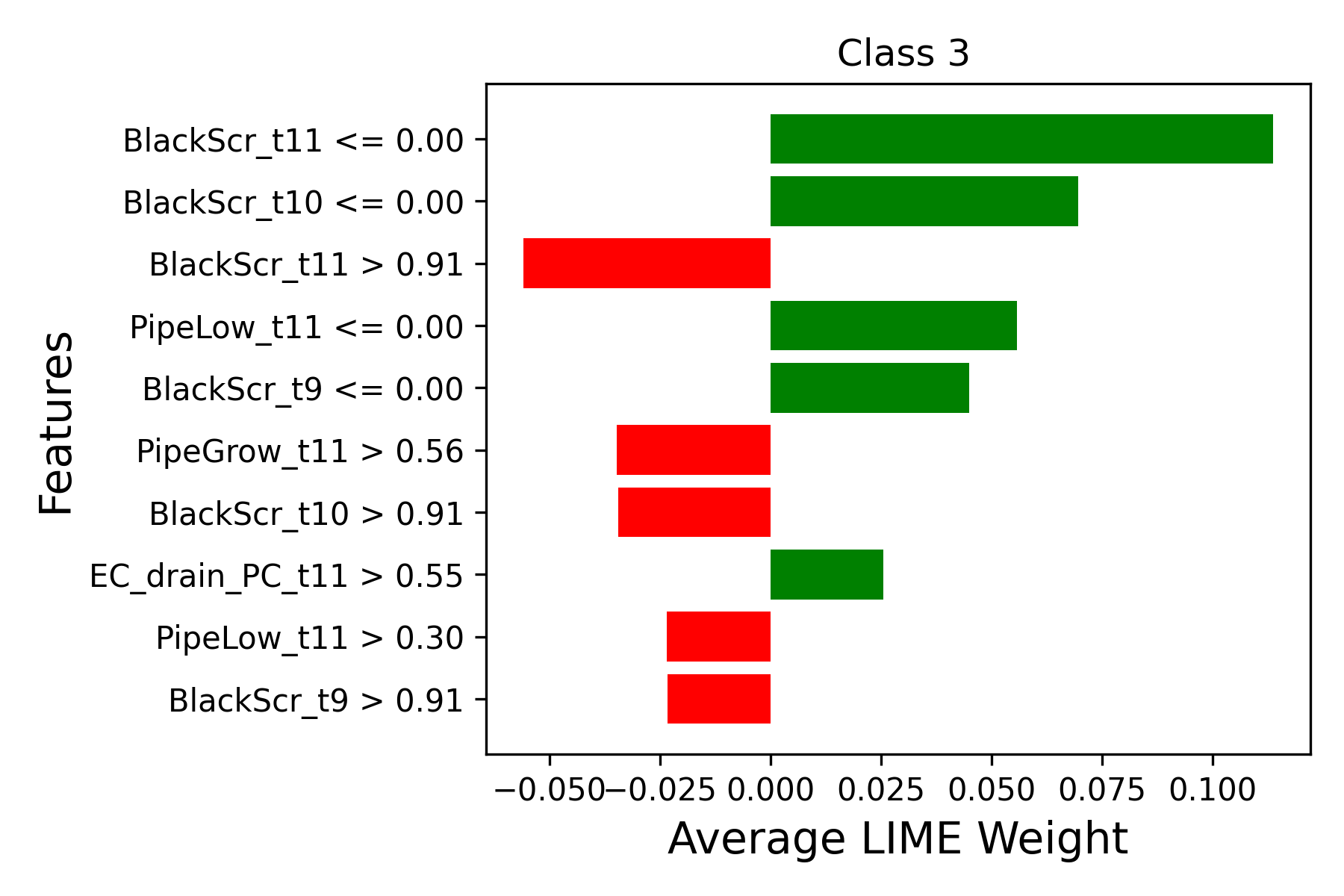}
        \subcaption{}
    \end{minipage}
    \hfill
    \begin{minipage}{0.32\textwidth}
        \centering
        \includegraphics[width=\linewidth, trim={0 0 0 0.8cm}, clip]{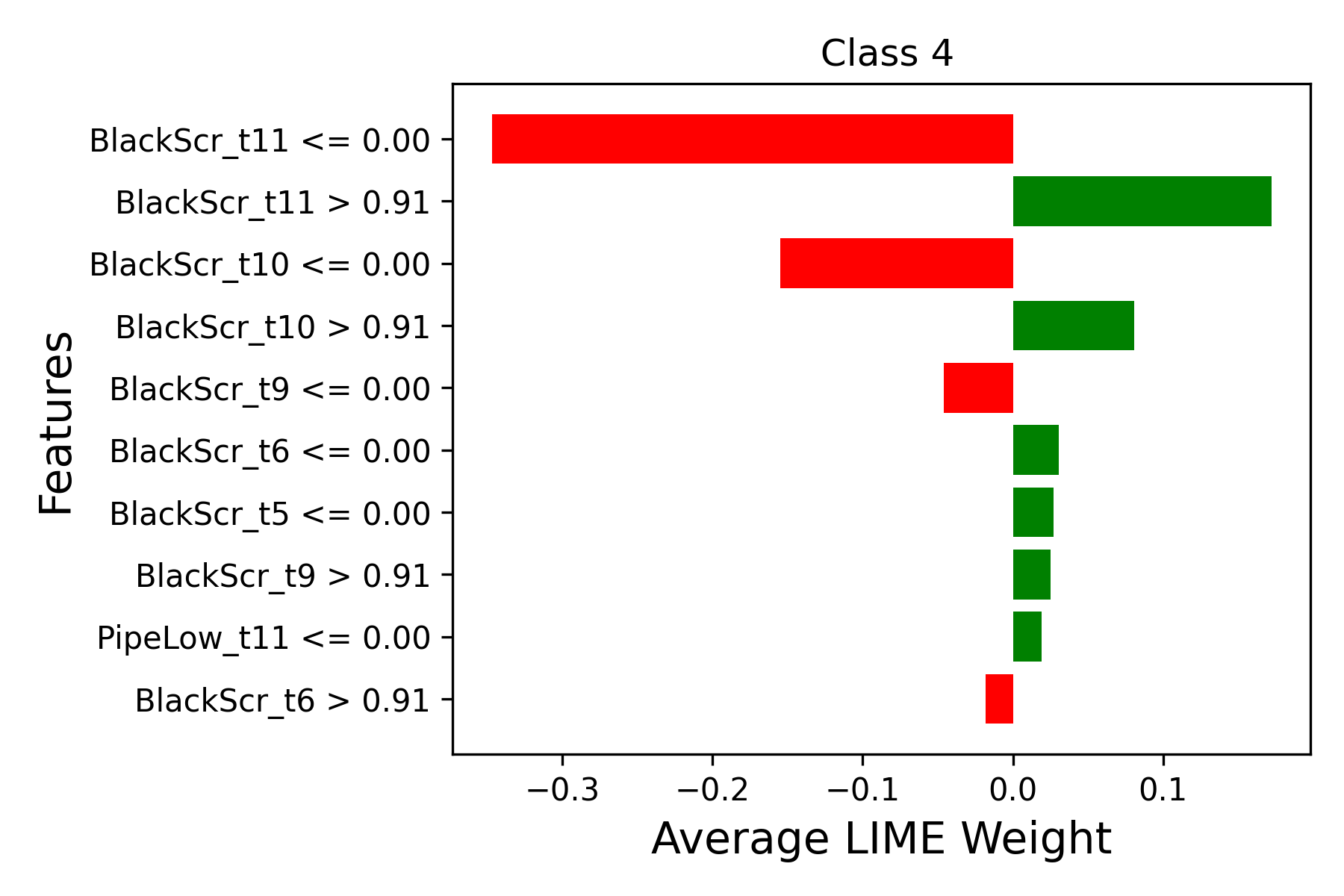}
        \subcaption{}
    \end{minipage}
    \hfill
    \begin{minipage}{0.32\textwidth}
        \centering
        \includegraphics[width=\linewidth, trim={0 0 0 0.8cm}, clip]{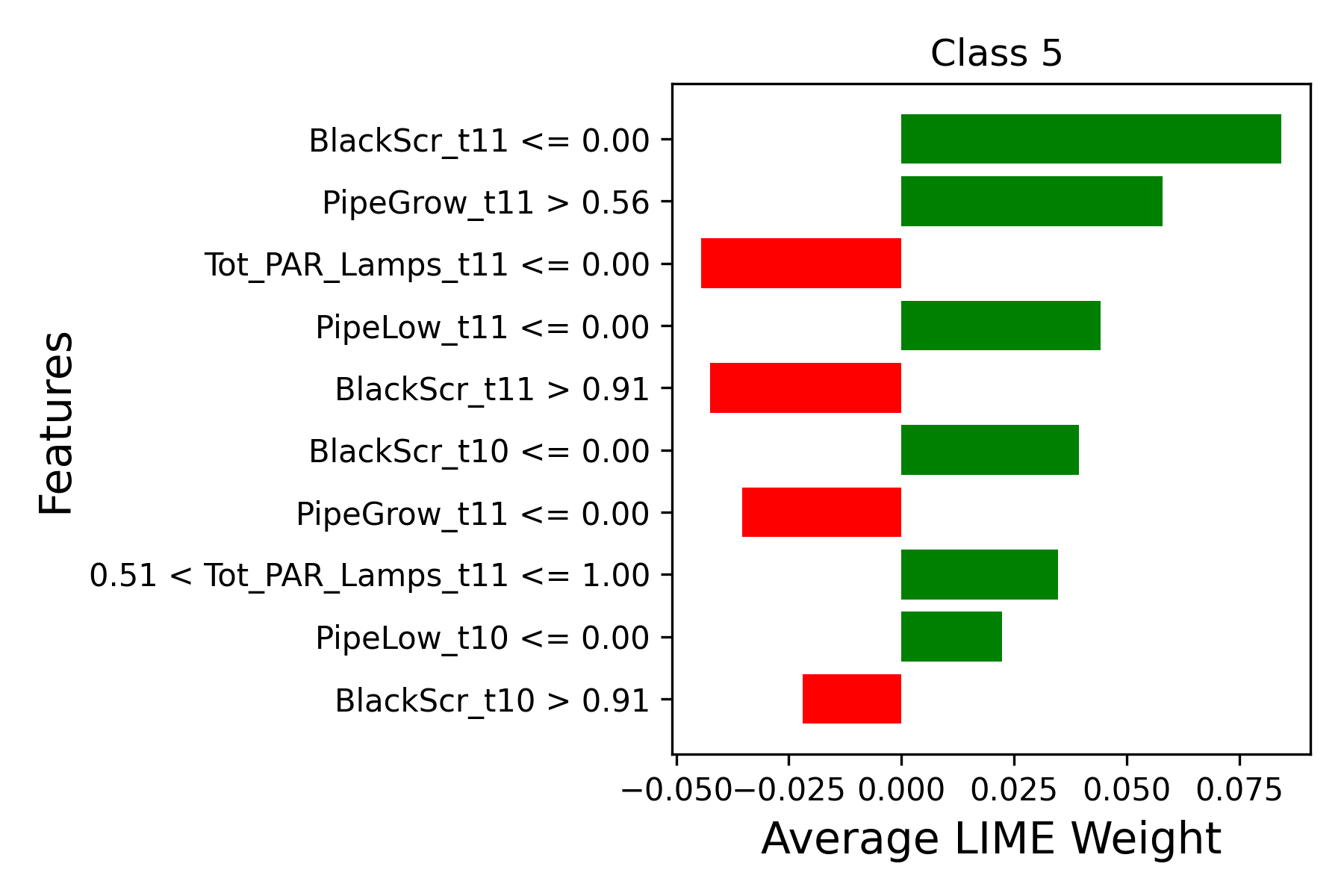}
        \subcaption{}
    \end{minipage}

    \caption{Top-10 influential features per actuator class (1–6) based on LIME explanations.}

    \label{fig:lime_explanations}
\end{figure*}

\subsection{Greenhouse Control}
To facilitate automation in greenhouse control, unsupervised machine learning \cite{kernel-kmeans} was used to identify clusters among 16 actuators and label them as six distinct classes (actuator setting class) to be activated for each input instance (sensor values). Several cluster counts were empirically tested, and six clusters were assumed to be optimal. The activation settings are described in Table \ref{tab:setting_classes}.

\subsection{Model Performance}

The model demonstrates stable convergence over 100 epochs, with training accuracy reaching 97\% and validation accuracy approaching 95\% (Figure~\ref{fig:training_accuracy}). The loss curves in Figure~\ref{fig:training_loss} show a consistent decrease for both the training and validation sets, further supporting the robustness of the model. The confusion matrix (Figure~\ref{fig:confusion_matrix}) illustrates a high classification accuracy across all six actuator classes, particularly for the dominant categories (3, 4, and 5). Misclassifications are distributed across both adjacent and non-adjacent classes, reflecting the complexity and variability of sensor conditions and control responses.

Overall, the model demonstrates good performance with class-wise F1-Score exceeding 0.88, confirming its effectiveness in learning complex actuator behaviour under varying environmental states.

Although high performance is crucial, it must be coupled with interpretability to ensure trust in the model deployment in IoRT-driven smart agriculture. Therefore, in subsequent sections, explainability techniques (VSN-based, LIME, and SHAP) are integrated to uncover the sensor-signal contributions driving these actuator decisions, enabling stakeholders to validate and fine-tune model behavior in real-time settings.

\subsection{Explainability of TFT-based Predictions}
SHAP provides global explanations to help understand the model's overall inference behaviour, and LIME helps to understand the model's reasoning for individual instances at the local level. Thus, both interpretations are compared with the model's VSN feature importance score to evaluate agreement on the feature importance among all XAI techniques.
\subsubsection{Global Feature Importance: VSN and SHAP}
The VSN, an intrinsic component of the TFT model, highlights which input variables were dynamically attended to during the temporal modelling process. As shown in Figure~\ref{fig:vsn}, features such as BlackScr, PipeGrow, and EC\_drain\_PC are most influential, indicating the model's sensitivity to shading conditions, heating infrastructure, and electrical conductivity in drainage—critical parameters for actuator state regulation.

In addition, the global SHAP analysis (Figure~\ref{fig:shap}) reveals Tair (air temperature) and Tot\_PAR (total photosynthetically active radiation) as the main contributors to actuator decisions. In particular, PipeGrow and Rhair also emerged as influential, validating the relevance of thermal and humidity conditions in greenhouse climate control. The convergence of VSN and SHAP insights confirms the robustness of feature attribution across independent interpretation methods.

\subsubsection{Local explanation with LIME}
LIME is applied to each of the six actuator classes derived through clustering to gain actionable insights into the sensor-to-actuator decision mappings at an instance level. These classes represent bundled actuator strategies commonly observed in automated greenhouse systems. For \textbf{Class 0}, comprising actuators such as \texttt{t\_rail\_min\_vip} and \texttt{t\_ventwind\_vip}, LIME identifies low values of \texttt{PipeLow} and \texttt{BlackScr} at recent time steps as key negative contributors, indicating a passive climate strategy with minimal heating and full shading, typically used during stable or nighttime conditions. \textbf{Class 1}, containing lighting-related actuators (\texttt{assim\_vip}, \texttt{int\_farred\_vip}, \texttt{int\_white\_vip}), is influenced positively by \texttt{AssimLight}, \texttt{Tot\_PAR\_Lamps}, and \texttt{PipeGrow}, reflecting the need to activate artificial lighting and heating under low radiation or photosynthetic demand. In \textbf{Class 2}, which includes \texttt{co2\_vip}, \texttt{int\_red\_vip}, \texttt{scr\_enrg\_vip}, and \texttt{window\_pos\_lee\_vip}, LIME highlightes the combined effects of \texttt{VentLee}, \texttt{CO2air}, and \texttt{BlackScr}, suggesting that this actuator setting aims to balance ventilation, gas exchange, and light shielding for optimal photosynthesis. \textbf{Class 3}, consisting of \texttt{dx\_vip}, \texttt{t\_heat\_vip}, \texttt{water\_sup\_intervals\_vip\_min}, and \texttt{t\_ventlee\_vip}, exhibits high sensitivity to \texttt{Tair}, \texttt{HumDef}, and \texttt{PipeGrow}, indicating its use in correcting abrupt climate deviations through aggressive heating and watering strategies. \textbf{Class 4}, which focuses solely on the \texttt{scr\_blck\_vip} (blackout curtain), is heavily driven by time-lagged values of \texttt{BlackScr}, confirming its tight coupling with light insulation needs during specific growth phases. Finally, \textbf{Class 5}, comprising spectral light (\texttt{int\_blue\_vip}) and thermal controls (\texttt{t\_grow\_min\_vip}), is predominantly influenced by \texttt{Tot\_PAR\_Lamps}, \texttt{PipeGrow}, and \texttt{EC\_drain\_PC}, representing fine-tuned interventions for maintaining growth-stage-specific conditions under fluctuating radiation and nutrient levels. These results confirm that the model captures both the functional interdependence of sensors and the agronomic rationale behind actuator coordination.

A comprehensive explanation using VSN, SHAP, and LIME offers rich interpretability and transparency to the greenhouse operational team to understand resource usage, which helps in planning the IoRTs. VSN and SHAP provide consistent, system-level attribution by identifying globally dominant environmental signals, while LIME dissects temporal and feature interactions at the actuator-class level. Integrating global (model-based and model-agnostic) and local explanations enhances stakeholder trust while facilitating actionable model debugging, ensuring policy compliance, and supporting domain-specific refinement of autonomous smart agriculture systems. The trained XAI system aims to replicate and enhance historical control strategies by learning sensor-actuator patterns from data. It provides real-time transparency and acts as a supervisory agent, generating actuator settings that can be interpreted and validated instantaneously.

\section{Conclusion and future work}
The integration of IoRT into smart farming offers a promising pathway toward scalable, climate-resilient, and sustainable food production. However, for such autonomous systems to gain widespread adoption, they must incorporate transparency and interpretability, enabling domain experts and stakeholders to understand, validate, and collaborate with AI-driven decision processes. 
The model achieves a classification accuracy of 95\% and is complemented by model-specific and model-agnostic XAI techniques, i.e., VSN, SHAP, and LIME to provide global and local explanations of how environmental factors influence actuator control decisions. This work proposes an explainable framework for autonomous greenhouse control that integrates an TFT model based on time-series sensor data. The VSN module identifies key shading, heating, and nutrient flow features. SHAP adds a broader environmental context by highlighting influential variables, such as temperature, light, and humidity. LIME reveals class-specific patterns in the importance of features in different actuator settings. Together, these methods provide complementary insights into the model’s global and local decision-making.

Future work aims to focus on edge deployment of the proposed framework to optimize resource use and crop yield, while incorporating active learning frameworks for adaptive control.

\section{Acknowledgements}
This work is funded by TU RISE with the support of the Government of Ireland and the EU through the ERDF Southern, Eastern \& Midland Regional Programme 2021–27 and the Northern \& Western Regional Programme 2021–27.

\bibliographystyle{IEEEtran}
\bibliography{biblio}

\end{document}